\documentclass[conference]{IEEEtran}
\IEEEoverridecommandlockouts
\usepackage{cite}
\usepackage{amsmath,amssymb,amsfonts}
\usepackage{algorithmic}
\usepackage{graphicx}
\usepackage{textcomp}
\usepackage{xcolor}
\usepackage{hyperref}

\usepackage{lipsum}
\usepackage{amsmath}
\usepackage{amssymb}
\usepackage{graphicx}
\usepackage{algorithmic}
\usepackage{adjustbox}
\usepackage{bm}
\usepackage{resizegather}
\usepackage[ruled, linesnumbered, longend]{algorithm2e}
\def\BibTeX{{\rm B\kern-.05em{\sc i\kern-.025em b}\kern-.08em
    T\kern-.1667em\lower.7ex\hbox{E}\kern-.125emX}}
\begin{document}

\title{\textbf{Sentiment Analysis of Movie Reviews Using BERT}}




\author{\IEEEauthorblockN{Gibson Nkhata}
\IEEEauthorblockA{\textit{Department of Computer Science \& Computer Engineering} \\
\textit{University of Arkansas}\\
Fayetteville, AR 72701, USA \\
Email: gnkhata@uark.edu}
\and
\IEEEauthorblockN{Usman Anjum, Justin Zhan}
\IEEEauthorblockA{\textit{Department of Computer Science} \\
\textit{University of Cincinnati}\\
Cincinnati, OH 45221, USA \\
Email: anjumun@ucmail.uc.edu, zhanjt@ucmail.uc.edu}

}

\maketitle

\begin{abstract}
Sentiment Analysis (SA) or opinion mining is  analysis of emotions and opinions from any kind of text. SA helps in tracking people’s viewpoints, and
it is an important factor when
it comes to social media monitoring, product and brand recognition, customer satisfaction,
customer loyalty, advertising and promotion’s success, and product acceptance. That is why SA is one of the active research areas in Natural Language Processing (NLP). SA is applied on data sourced from various media platforms to mine sentiment knowledge from them. Various approaches have been deployed in the literature to solve the problem. Most techniques devise complex and sophisticated frameworks in order to attain optimal accuracy.  
This work aims to fine-tune  Bidirectional Encoder Representations from Transformers (BERT) with Bidirectional Long Short-Term Memory (BiLSTM) for movie reviews sentiment analysis and still provide better accuracy than the State-of-The-Art (SOTA) methods. The paper also shows how sentiment analysis can be applied, if someone wants to recommend  a certain movie, for example, by computing overall polarity of its sentiments predicted by the model. That is, our proposed method serves as an upper-bound baseline in prediction of a predominant reaction to a movie. To compute overall polarity, a heuristic algorithm is applied  to BERT-BiLSTM output vector. 
Our model can be extended to three-class, four-class, or any fine-grained classification, and apply overall polarity computation again. This is  intended to be exploited in future work. 
 
\end{abstract}


\section{Introduction}



Sentiment analysis aims to determine the polarity of emotions like happiness, sorrow, grief, hatred, anger and affection and opinions from text, reviews, and posts which are available in 
media platforms~\cite{baid2017sentiment}. With the emergence of various social media platforms, vast amount of data are contained and various information, e.g., education, health, entertainment, etc, is shared in these online forums 
everyday across the globe. Therefore, there have been advances in many Natural Language Processing (NLP) tasks in conjunction with machine or deep learning techniques that automatically mine knowledge from the data sourced from these repositories~\cite{socher2013recursive}.
As an NLP task, sentiment analysis helps in tracking people’s viewpoints. For example, it is a powerful marketing tool that enables product managers to understand customer emotions in their various marketing campaigns. It is an important factor when it comes to social media monitoring, product and brand recognition, customer satisfaction, customer loyalty, advertising and promotion’s success, and product acceptance. Sentiment analysis is among the most popular and valuable tasks in the field of NLP~\cite{mesnil2014ensemble}. 

Movie reviews is an important approach to gauge the performance of a particular movie. Whereas providing a numerical or stars rating to a movie quantitatively tells us about the success or failure of a movie, a collection of movie reviews is what gives us a deeper qualitative insight on different aspects of the movie. A textual movie review tells us about the strengths and weaknesses of the movie and deeper analysis of a movie review tells if the movie generally satisfies the reviewer.

Movie Reviews Sentiment Analysis is being worked on in this study because movie reviews have standard benchmark datasets, where salient and qualitative works have been published on. In fact, most of these reviews are crawled from the social media platforms\cite{fang2015sentiment}.


 Bidirectional Encoder Representations
from Transformers (BERT) is a popular pre-trained language representation model and has proven to perform well on many NLP tasks like question answering, named entity recognition, and text classification \cite{devlin2018bert}. BERT has been used in many works for sentiment analysis, like  in \cite{munikar2019fine}. However, regarding the capabilities of BERT, the performance is not satisfactory.

In this paper, BERT is fine-tuned for sentiment analysis on movie reviews and provide optimal accuracy that surpass accuracy of State-Of-The Art (SOTA) models. Our focus is on polarity  classification on a 2-point scale. Polarity classification classifies a text as containing either a negative or positive sentiment. 

 BERT has proven to be satisfactory in many NLP downstream tasks. BERT has been used in information retrieval in \cite{guo2020detext} to build an efficient ranking model for industry use cases. The pre-trained language model was also successfully utilised in \cite{liu2019fine} for extractive summarization of text and used for question answering with satisfactory results in \cite{he2020infusing}. Yang et al. \cite{yang2019data} efficiently applied the model in data augmentation resulting in optimal results.  

Fine-tuning is a common technique for transfer learning. The target model copies all model designs with their parameters from the source model except the output layer and fine-tunes these parameters based on the target dataset. The main benefit of fine-tuning is no need of training the entire model from scratch, and only the output layer of the target model needs to be trained. Hence, BERT is being fine-tuned in this work by coupling with Bidirectional Long Short-Term Memory (BiLSTM) and train the resulting model on movie reviews sentiment analysis benchmark datasets. BiLSTM processes input features bidirectionally~\cite{devlin2018bert}, which helps in improving target model generalisation. Therefore, our fine-tuning approach is called BERT+BiLSTM-SA, where SA stands for Sentiment Analysis.

Finally, how results of sentiment analysis can be applied is shown. If someone wants to recommend  a certain movie, for example, by computing overall polarity of its reviews sentiments predicted by the model. That is, the proposed method serves as an upper-bound baseline in prediction of the polarity of predominant reaction to a movie. To compute overall polarity, a heuristic algorithm adopted from \cite{arasteh2021will} is applied to BERT-BiLSTM+SA output vector. The algorithm in their paper is applied on the output vector of three class classification on twitter dataset by LSTM, whereas our approach customises the algorithm for binary classification output vector from BERT+BiLSTM-SA. 

This work is divided into two main components. First, fine-tuning BERT with BiLSTM and use the resulting model on binary sentiment polarity classification. Second, using the results of sentiment classification  in computation of a predominant sentiment polarity.

\subsubsection{Our contributions}
Our contributions in this work are:
\begin{itemize}
\item 
Fine-tuning BERT  by coupling with  BiLSTM  for  polarity classification on well known benchmark datasets and achieve accuracy that beats SOTA models.

\item
Computing overall polarity  of predicted reviews from BERT+LSTM-SA output vector.
\item
Comparing our experimental outcomes with results obtained from other studies, including SOTA models, on benchmark datasets. 
\end{itemize}
This paper is organised as follows. Section \ref{RelatedWork} describes related work, Section \ref{Methodology} describes the methodology, Section \ref{Experiments} discusses experiments and results, and last, Section \ref{conclusion} gives conclusion and talks about future work. The code for this project is available \cite{bcode} 
to enable wider adoption.

\section{Related work} \label{RelatedWork}
A lot of work has been conducted in literature on movie reviews sentiment analysis. Starting with traditional machine learning, a step-by-step lexicon-based sentiment analysis using the R open-source software is presented in \cite{anandarajan2019sentiment}. In \cite{baid2017sentiment}, the authors implemented and compared traditional machine learning techniques like Naive Bayes (NB), K-Nearest Neighbours (KNN), and  Random Forests (RF) for sentiment analysis. Their results showed that Naive Bayes was the best classifier on the task. An ensemble generative approach for various machine learning approaches on sentiment analysis is used in \cite{mesnil2014ensemble}. KNN with the help of information gain technique was also used in \cite{daeli2020sentiment} on the task. In \cite{thongtan2019sentiment}, the authors proposed training document embeddings using cosine similarity, feature combination, and NB. KNN outperforms all other models in these works.

Deep learning approaches have also been implemented in movie reviews sentiment analysis. Recurrent Neural Network (RNN) and Convolutional Neural Network (CNN) architectures performances were explored for semantic analysis of movie reviews in \cite{shirani2014applications}. RNNs give satisfactory results, but they suffer from the problem of vanishing or exploding gradients when used with long sentences. Nonetheless, CNNs provide non-optimal accuracy on text classification. Coupled Oscillatory RNN (CoRNN), which is a time-discretization of a system of second-order ordinary differential equations, was proposed in \cite{rusch2020coupled} to mitigate the exploding and vanishing gradient problem, though the performance was still not convincing.  Bodapati et al.~\cite{bodapati2019sentiment} used LSTM on movie reviews sentiment analysis by investigating the impact of different hyper parameters like dropout, number of layers, and activation functions. Additionally, BiLSTM network for the task of text classification has also been applied via mixed objective function in \cite{singh2020revisiting}. BiLSTM achieved better results but at the expense of a very sophisticated architecture.  

BERT has also been previously applied to sentiment analysis. BERT was used on SST-2 movie reviews benchmark for sentiment analysis in \cite{munikar2019fine}. In \cite{sousa2019bert}, the authors used BERT for stock market sentiment analysis. BERT was also applied on target-dependent sentiment classification in \cite{gao2019target}. However, there is still room for improvement considering their results. 

Therefore, in this work, BERT is fine-tuned  by coupling with BiLSTM for sentiment analysis on a 2-point scale. Afterwards, an application of sentiment  analysis is shown by computing overall polarity of movie reviews, which can also be utilised in recommending a movie.

\section{Methodology} \label{Methodology}
Different techniques used in our work starting with description of sentiment analysis and BERT are covered in this section. Afterwards,  the section explains how BERT is fine-tuned with BiLSTM, elucidates how classification is applied in our work, describes overall polarity computation, and talks about the overview of the whole work.

\subsection{Sentiment analysis}
Sentiment analysis is a sub-domain of opinion mining, which aims at the extraction of emotions and opinions of people towards a particular topic from a structured, semi-structured, or unstructured textual data \cite{gadekallu2019application}.
In our context, the primary objective is to classify a review  as carrying either a positive or negative sentiment on a 2-point scale. 

\subsection{BERT}
BERT  was introduced by researchers from Google\cite{devlin2018bert} and focuses on pre-training deep bidirectional representations from unlabeled text by jointly conditioning on both left and right contexts in all layers of the model. As a result, BERT can be fine-tuned with an extra component on a downstream task like question answering and sentiment analysis. 

 There are two primary models of BERT, namely, BERT\textsubscript{BASE} and BERT\textsubscript{LARGE}.  BERT\textsubscript{BASE}, which has 12 layers, 768 hidden states, 12 attention heads, and 110M parameters is used in this work, whereas BERT\textsubscript{LARGE} has almost 2 times of each of these specifications. Actually, the uncased version of  BERT\textsubscript{BASE} known as \textit{bert-base-uncased}, which accepts tokens as lowercase, is adapted in this work. Because of multiple attention heads, BERT processes a sequence of input tokens in parallel, thereby improving the model generalization on the input sequence.

BERT has its own format for the input tokens. The first token of every sequence is denoted as [CLS]. The token corresponds to the last hidden layer, aggregates all the information in the input sequence, and is used for classification tasks.  Sentences are packed into a single input sequence and differentiated in two ways: using a special token [SEP] to separate them and adding a learned embedding to every token identifying a sentence where it belongs to. 

\begin{figure}[!h] 
\centering
\includegraphics[width=0.3\textwidth]{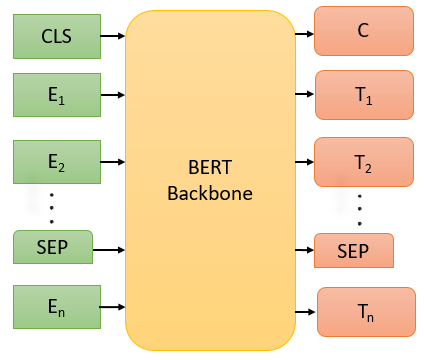}
\caption{Simplified diagram of BERT}
\label{fig:fig1}
\end{figure}

Figure \ref{fig:fig1} shows a simplified diagram of BERT. $E_n$ is an  input representation of a single token constructed by summing the corresponding token, segment, and position embeddings; BERT Backbone represents main processing performed by BERT; $T_n$ is a hidden state corresponding to token $E_n$; and $C$ is a hidden state corresponding to aggregate token $CLS$.

\subsection{Fine-tuning BERT with BiLSTM}
Since BERT is pretrained,  
there is no need of training the entire model from scratch. Hence, information is just needed to be transferred  from BERT  to the fine-tuning component and train the model for sentiment analysis. This saves training time of the resulting model. 

\begin{figure*}[!h]
\center
\includegraphics[width=0.9\textwidth]{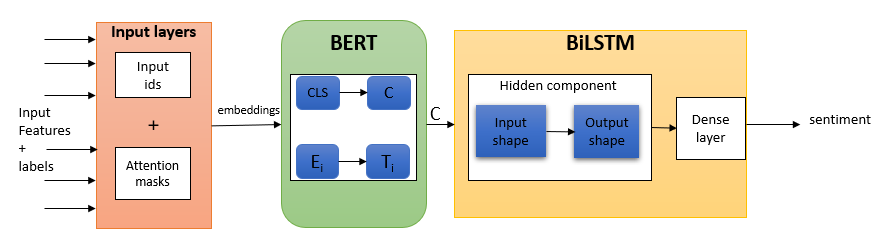}
\caption{Fine-tuning of the model} 
\label{fig:fig2}
\end{figure*}

Fine-tuning in this work is conducted as follows. After data preprocessing,  two input layers to BERT are built, where names of the layers need to match the input values. 
These input values are attention masks and input ids, as shown in Figure \ref{fig:fig2}. In other words, attention masks and input ids are input embeddings to the model. 

The input embeddings are propagated through BERT afterwards. Dimensionality of the embeddings depends on the input sequence length, batch-size and number of units in a layer. 
BiLSTM is then concatenated at the very end of BERT, and it includes a dense layer. Therefore, BiLSTM receives information from BERT and feeds it into its dense layer, which then predicts respective sentiments for the input features. BERT and BiLSTM shared same hyperparameters, and all hyperparameters are specified in Section \ref{Experiments} under experimental settings.

The fine-tuning part of our model is illustrated in Figure \ref{fig:fig2}. In the figure, \textit{input features} are tokens in a review text, \textit{input ids} identify a sentence that a token belongs to, and \textit{attention masks} are binary tensors indicating the position of the padded indices to a particular sequence so that the model does not attend to them. For the attention mask, 1 is used to indicate a value that should be attended to, while 0 is used to indicate a padded value. Padding helps in making sequences have same length when sentences have variable lengths, which is common in NLP. Therefore, padded information is not part of the input and should rarely be used in model generalization. 

The output from BERT has the same dimension as the input to BiLSTM.  $E_i$ and $T_i$  mean similar items as $E_n$ and $T_n$, respectively, in Figure \ref{fig:fig1}. BiLSTM has only one hidden component. 
Finally, there is a fully connected layer (dense layer) at the end, which has output dimension of batch-size by 1, since binary classification is being worked on here. The dense layer predicts a sentiment polarity.

 Weights from first layers of BERT are frozen  so that our focus dwells on the last layers close to the fine-tuning component. These layers contain trainable weights, which are updated to minimize the loss during training of the model on the downstream task of sentiment analysis. 


\subsection{Classification}
In this work, BERT is  fine-tuned on  polarity classification or binary classification of sentiment analysis. 
Hence, classification in this context is defined as follows. Given a movie review \textit{R}, classify it as carrying either a positive sentiment or a negative sentiment.

\subsection{Overall polarity}
 Overall polarity is defined as follows. Given an output vector from BERT+BiLSTM-SA containing sentiment labels of \textit{N} reviews, find the dominating sentiment polarity in the vector. To compute the overall polarity of reviews, the output vector of BERT+BiLSTM-SA is fed into the heuristic algorithm. First, number of labels for each class is counted in the output vector. Then, the computation is conducted as shown in Algorithm \ref{alg1}. While Arasteh et al. \cite{arasteh2021will} used Algorithm \ref{alg2} to compute overall polarity from the results of three-class classification on twitter replies by LSTM for twitter sentiment analysis, 
 this work adapts the algorithm to compute overall polarity from output vector of binary classification by BERT+BiLSTM-SA.

 Algorithm \ref{alg2} shows how overall polarity is computed from output vector of three-class classification. The overall polarity is first considered to be neutral if the proportion of neutral samples is at least higher than an empirically set threshold, which was set to be $85\%$ of the total predictions. The reason being that  most  samples are generally expected to be neutral, for not every text can be expected to carry a positive or a negative polarity. Then, negative and positive sentiments are considered in the  output vector. Again, there is usually no exclusively positive or negative text sample, that is why a positive overall sentiment is assigned if there is at least 1.5 times as many negative reviews as positive reviews, and vice versa. Meaning that the size of the dominating class must be at least higher for all the reviews to carry its polarity.  Lastly, a neutral sentiment is given when the total numbers of positive and negative reviews are close to each other, implying nonexistence of dominance between the two sentiments in the reviews. All constant values in the algorithm were set depending on the various empirical observations in the experiments.

\begin{algorithm}[t!] \label{alg2}
\caption{ Computing overall polarity from three-class classification output vector as done in \cite{arasteh2021will}.}
\KwResult{Dominating sentiment for all reviews.} 

\uIf{$\#$total neutral reviews $> 85\%$ of the total reviews}{\textit{overall polarity} $\gets$ \textit{neutral}\;}
\uElse{
    \uIf{$\#$total positive reviews $> $1.5 $\times$ \# of total negative reviews}{\textit{overall polarity} $\gets$ \textit{positive}\;}
    
    \uElseIf{$\#$total negative reviews $> $1.5 $\times$ \# of total positive reviews}{\textit{overall polarity} $\gets$ \textit{negative}\;}
    \uElse{\textit{overall polarity} $\gets$ \textit{neutral}\;}
}
\end{algorithm}

 Algorithm \ref{alg1} is derived from Algorithm \ref{alg2} by us. Depending on various empirical observations in our experiments, a different threshold coefficient is used when multiplying. Additionally, only proportions of two sentiments are being compared here. Although there is not a neutral sentiment polarity in the output vector of our model, it is introduced for the overall polarity computation if the output of the first two conditions in Algorithm \ref{alg1} is false, implying a tie between positive and negative reviews.

A naive approach to computing the overall polarity would be just counting the number of labels for each class in BERT+BiLSTM-SA output vector and assign the overall polarity depending on majority class. However, the overall polarity computed from this approach cannot represent a good dominating majority class. For example, assume there is 102 predictions of movie reviews for one movie, and the output vector has 50 positive reviews and 52 negative reviews, the overall polarity will be negative. However, if this is applied in recommending a movie, for example, a difference of 2 is not optimal to determine the dominating polarity of all the movie reviews and conclude that the movie is not interesting. Additionally, the level of positiveness or negativity  is different for every review in the datasets. As a result, the formulations in Algorithm \ref{alg1} are used so that either class in the output vector must contain a bigger proportion to assign its label as the output, otherwise the overall polarity becomes neutral.

The overall polarity computation technique can be used to recommend a movie to a person disregarding subjectivity factors. First,  all the reviews pertaining to the movie in question must be gathered, followed by using  BERT+BiLSTM-SA to predict the associated sentiments polarities for the reviews. Then,  the algorithm can be used to compute the dominating sentiment polarity to see if most people are in favor of the movie or not. Last, the movie can be recommended  if the overall polarity is positive.

\DontPrintSemicolon
\begin{algorithm} [h!] \label{alg1}
\caption{ Computing overall polarity from binary classification output vector}
\KwResult{Dominating sentiment for all reviews.} 

\uIf{$\#$total positive reviews $> $1.2 $\times$ \# of total negative reviews}{\textit{overall polarity} $\gets$ \textit{positive}\;}

\uElseIf{$\#$total negative reviews $> $1.2 $\times$ \# of total positive reviews}{\textit{overall polarity} $\gets$ \textit{negative}\;}

\uElse{\textit{overall polarity} $\gets$ \textit{neutral}\;}
\end{algorithm}

\subsection{Overview of our work}
\begin{figure*}[h!]
\center
\includegraphics[width=0.7\textwidth]{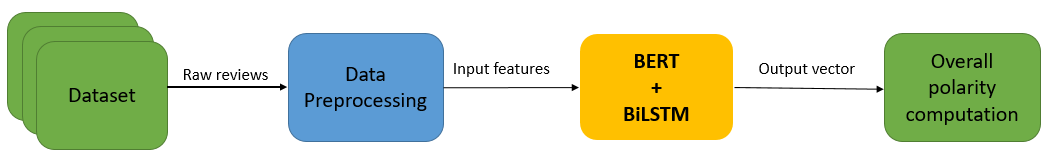}
\caption{Overview of our work} 
\label{fig:fig3}
\end{figure*}
Figure \ref{fig:fig3} portrays the overview of our work. In a nutshell, the work starts with  preprocessing raw text data into features that can be understood by BERT and feed the features into BERT (BERT\textsubscript{BASE} is the same as BERT in the diagram) and BiLSTM  through the fine-tuning layer, which specifies the hyperparameters that BERT+BiLSTM should use. Lastly, an output vector from  BERT+BiLSTM predictions is used to compute overall polarity.

\section{Experiments} \label{Experiments}
This section starts with an explanation of datasets that are used in the experiments followed by data preprocessing. Afterwards, it describes experimental settings, explains evaluation metrics used in experiments, and discusses experimental results.

\subsection{Datasets}
Datasets used in the experiments consist of reviews annotated for sentiment analysis on a 2-point scale. Following is the description of the datasets used and how some of them were changed to suit experimental settings. Table \ref{tab3} shows statistics of the datasets used in the experiments.

\begin{table}[!h]
\caption{Statistics of the datasets divided into training and test sets }
\begin{center}
\begin{tabular}{|c|c|c|c|c|}
\hline
\textbf{Dataset}&\multicolumn{2}{|c|}{\textbf{Train samples}} &\multicolumn{2}{|c|}{\textbf{Test samples}}\\
\cline{2-5} 
\textbf{} &POSITIVE & NEGATIVE&POSITIVE  &NEGATIVE  \\
\hline
IMDb &12500 &12500 &12500 &12500 \\
\hline
MR &4264 &4265 &1067 &1066\\
\hline
SST-2  & 4300 & 4244 &886 &1116\\
\hline
Amazon&239660  &37056 &59949 &9231\\
\hline
\end{tabular}
\label{tab3}
\end{center}
\end{table}

\subsubsection{IMDb movie reviews} Introduced in \cite{maas2011learning}, IMDb movie reviews dataset is a binary sentiment analysis dataset consisting of 50,000 reviews from the Internet Movie Database (IMDb), where the name comes from. It comprises equal number of negative and positive reviews.

\subsubsection{SST-2} The SST (Stanford Sentiment Treebank) is a corpus with fully labeled parse trees that allows for a complete analysis of the compositional effects of sentiment in language. The corpus is based on the dataset introduced in \cite{socher2013recursive}, and it consists of 11,855 single sentences extracted from movie reviews. It was parsed with the Stanford parser and includes a total of 215,154 unique phrases from those parse trees, each annotated by three human judges. In our work, SST-2 version of the dataset is used. SST-2 is designed for binary classification and consists of 11855 movie reviews.

\subsubsection{MR Movie Reviews} MR Movie Reviews dataset comprises collections of movie review documents labeled with respect to their overall sentiment polarity, positive or negative, or subjective rating, for example two and a half stars, and sentences labeled with respect to their subjectivity status, subjective or objective, or polarity. In this paper, the version introduced in \cite{pang2005seeing}, which consists of 5331 positive and 5331 negative processed reviews is used.

\subsubsection{Amazon Product Data dataset} This dataset contains product reviews and metadata from Amazon, including 142.8 million reviews spanning May, 1996 through July, 2014. The dataset includes reviews, product metadata, and links. It was introduced in \cite{fang2015sentiment} for sentiment analysis using product review data, and in \cite{he2016ups} to build a recommender system in collaborative filtering setting on amazon products. In our work, only video reviews are focused on. There is a total of 345896 video reviews samples. The dataset originally contained labels with scores from 1 to 5 corresponding to polarity strength variation from negative to positive. The dataset was then prepared for binary classification by replacing 1 and 2 scores with a negative label and 4 and 5 scores with a positive label, and score 3, which represents a neutral class, was discarded as in SST-2\cite{socher2013recursive}.

\subsection{Data preprocessing}
The needed data preprocessing steps require to transform the input data into a format that BERT can understand. This involves carrying out two primary data preprocessing steps. 

First, creating input examples using the constructor provided in the BERT library. The constructor accepts three main parameters, which are \textit{text\_a}, \textit{text\_b}, and \textit{label}. \textit{text\_a} is the text that  the model must classify, which in this case, is the collection of movie reviews without their associated labels. \textit{text\_b} is used if a model is being trained to understand the relationship between sentences, for example sentence translation and question answering.  The previous scenario hardly applies in our work, so \textit{text\_b} is just left  blank. \textit{label}  has labels of input features. In our case, \textit{label} implies sentiment polarity of every movie review, which can be negative or positive. Refer to BERT original paper \cite{devlin2018bert}  for more details about this step and even the next step.

Last, the following preprocessing steps are conducted.
\begin{itemize}
    \item Lowercase text, since the lowercase version of BERT\textsubscript{BASE} is being used.
    \item Tokenize all sentences in the reviews. For example, "this is a very fantastic movie" to "this", "is", "a", "very", "fantastic", "movie".
    \item Break words into word pieces. That is  "interesting" to "interest" and "\#\#ing".
    \item Map words to indexes using a vocab file that is provided by BERT.
    \item Adding special tokens: [CLS] and [SEP], which are  used for aggregating information of the entire review through the model and separating sentences respectively.
    \item Append index and segment tokens to each input to track a sentence which a specific token belongs to.
\end{itemize}
The output of the tokenizer after these steps are \textit{input ids} and \textit{attention masks}. These are then taken as inputs to our model in addition to the reviews labels.

\subsection{Experimental settings}
Many simulations were carried on the datasets to find optimal hyperparameters for the model. As a result, optimal results from the experiments were obtained by the following settings.
256 input sequence length (\textit{K}), adam optimizer, 3e-5 learning rate, 1e-08 epsilon, and sparse categorical cross entropy loss. The model was trained  for 10 epochs and repeated steps for each batch. These hyperparameters were cordially fine-tuned regarding both BERT and BiLSTM, and overfitting was noticed when increasing the number of epochs for the model. 

\subsection{Evaluation metrics}
Accuracy was used to evaluate the performance of our model and compare it with other models. Accuracy metric is adopted because it is greatly applied in most works \cite{wang2021entailment, jiang2019smart, munikar2019fine, chen2022dual, chilkuri2021parallelizing}. Therefore, this adoption makes our work to be consistent with other works that are being compared against. Accuracy is defined as follows:

\begin{equation}
accuracy = \frac{\text{\textit{number of correct predictions}}}{\text{\textit{total number of predictions}}}\times 100
\end{equation}

\subsection{Results} \label{results}

Table \ref{tab1} presents accuracy comparisons between our model, BERT+BiLSTM-SA, and other models on all datasets. BERT+BiLSTM-SA outperforms other models on all the datasets, thereby achieving new SOTA accuracy on these benchmark datasets. The best accuracy of 98.76\% accuracy is obtained on Amazon dataset.

\begin{table}[!h]
\caption{Accuracy (\%) Comparisons  of Models on Benchmark Datasets for Binary Classification}
\centering
\begin{tabular}{|c|c|c|c|c|}
\hline
\textbf{Model name}&\multicolumn{4}{|c|}{\textbf{Dataset}} \\
\cline{2-5} 
\textbf{} & \textbf{\textit{IMDb-2}}& \textbf{\textit{MR}}& \textbf{\textit{SST-2}} & \textbf{\textit{amazon-2}} \\
\hline
RNN-Capsule \cite{wang2018sentiment} &84.12& 83.80&82.77&82.68 \\
\hline
coRNN \cite{semwal2018practitioners} & 87.4 & 87.11 & 88.97&89.32\\
\hline
TL-CNN \cite{semwal2018practitioners} & 87.70 & 81.5 & 87.70&88.12\\
\hline
Modified LMU \cite{chilkuri2021parallelizing} & 93.20&93.15&93.10&93.67\\
\hline
DualCL \cite{chen2022dual} &- & 94.31& 94.91 &94.98\\
\hline
L Mixed \cite{sachan2019revisiting} &95.68 &95.72&-&95.81 \\
\hline
EFL \cite{wang2021entailment} & 96.10 & 96.90 & 96.90 &96.91\\
\hline
NB-weighted-BON \cite{thongtan2019sentiment} & 97.40&-&96.55&97.55\\
+dv-cosine  & &&&\\
\hline
SMART-RoBERTa  \cite{jiang2019smart} & 96.34& 97.5 &96.61&-\\

 Large  & & &&\\
\hline
\textbf{Ours }& \textbf{97.67}&\textbf{97.88} & \textbf{97.62} & \textbf{98.76}\\
\hline
\multicolumn{5}{l}{}
\label{tab1}
\end{tabular}
\end{table}

Table \ref{tab4} shows results of ablation studies to see the impact of each component in the model. It can be seen that both BERT and BilLSTM separately give lower accuracy on the predictions compared against BERT+BiLSTM-SA. Therefore, the coupling  of the two tools enhances model generalization. 

\begin{table}[!h]
\caption{Results of Ablation Study }
\centering
\begin{tabular}{|c|c|c|c|c|}
\hline
\textbf{Model name}&\multicolumn{4}{|c|}{\textbf{Dataset}} \\
\cline{2-5} 
\textbf{} & \textbf{\textit{IMDb-2}}& \textbf{\textit{MR}}& \textbf{\textit{SST-2}} & \textbf{\textit{amazon-2}} \\
\hline
BiLSTM  &90.42& 90.5&91.12&92.18 \\
\hline
BERT  & 93.81 & 94.29 &93.55.97&94.78\\
\hline
\textbf{BERT+BiLSTM-SA }& \textbf{97.67}&\textbf{97.88} & \textbf{97.62} & \textbf{98.76}\\
\hline
\multicolumn{5}{l}{}
\label{tab4}
\end{tabular}
\end{table}

The discussion of results is finished by talking about the overall polarity computation on all datasets by  BERT+BiLSTM-SA. Table \ref{tab2}  presents the overall polarity computed from all the datasets. \textbf{Original overall polarity} is known before input embeddings are fed into BERT+BiLSTM-SA for prediction, while \textbf{Computed overall polarity} is computed and  known after BERT+BiLSTM-SA has made predictions on the reviews.  The table shows that the \textbf{Computed overall polarity} is the same as the \textbf{Original overall polarity} for all the datasets. The \textbf{Original overall polarity} is calculated by counting the number of samples of each label in the input  and use the result in the heuristic algorithm. That is, using original proportion of each class in the input before the model has made predictions on the reviews. The output was then used to verify the \textbf{Computed overall polarity}. The \textbf{Computed overall polarity} is computed similarly, but predictions are considered. Therefore, without loss of generality, there is confidence that the model predicts the expected sentiment on a given review and the heuristic algorithm computes accurate overall polarity from the model, regarding each dataset.

\begin{table}[h!]  
\caption{Overall Polarity Computation on All  Datasets}
\begin{center}
\begin{tabular}{|c|c|c|}
\hline
\textbf{Dataset}  & \textbf{Original overall polarity}  & \textbf{Computed overall polarity}\\
\hline
IMDb & Neutral & Neutral\\
\hline
MR reviews & Neutral & Neutral\\
\hline
SST-2& Neutral & Neutral\\
\hline
Amazon & Positive & Positive\\
\hline
\end{tabular}
\label{tab2}
\end{center}
\end{table}

\section{CONCLUSION} \label{conclusion}
Sentiment analysis is an active research domain in NLP. In this work, the existing domain knowledge of sentiment analysis is extended by providing another effective way of fine-tuning BERT to improve accuracy measure on movie reviews sentiment analysis and show how to compute an overall polarity of a collection of movie reviews sentiments predicted by a model, BERT+BiLSTM-SA, for example. To fine-tune BERT,  the technique of transfer learning was employed by coupling  BERT with BiLSTM.  BiLSTM, which had a dense layer, acted as a classifier on BERT final hidden states. The model was used for polarity classification and was experimented on IMDb, MR, SST-2, and Amazon datasets. It was also shown that sentiment analysis can be applied, if someone wants to recommend  a certain movie, for example, by computing overall polarity of its sentiments predicted by the model. That is, the proposed method serves as an upper-bound baseline in prediction 
of a predominant reaction to a movie.
Ablation studies also show that BERT and BiLSTM seperately provide non-optimal accuracy compared against BERT+BiLSTM-SA, implying coupling of the two tools is stronger for the model generalization.

To compute overall polarity, a heuristic algorithm is applied to BERT-BiLSTM+SA predictions. For all the datasets, it have been demonstrated that the original overall polarity is the same as the computed overall polarity. To the best of our knowledge, this is the first work to couple BERT with BiLSTM for sentiment classification task and use the model output vector to compute overall sentiment polarity. 
Our model is robust whereby it can be extended to three-class, four-class, or any fine-grained classification. This is intended to be explored in future work to prove the robustness of the model. 

Future work will additionally dwell on how to effectively apply accuracy improvement techniques to transformed BERT features despite loss of semantic information in them, exploring other pre-trained language models, and how different components of a sentence contribute to its sentiment prediction since this is information that is not generally explored by current works. 

\IEEEtriggeratref{13}
\bibliographystyle{IEEEtran}
\bibliography{mybib}

\end{document}